%% file: main.tex
\documentclass[10pt,twocolumn,letterpaper]{article}

\usepackage[pagenumbers]{cvpr} 

\usepackage{graphicx}
\usepackage{amsmath}
\usepackage{amssymb}
\usepackage{booktabs}
\usepackage{multirow}
\usepackage{enumerate}
\usepackage{enumitem}
\usepackage[toc,page]{appendix}
\usepackage[accsupp]{axessibility}

\usepackage[pagebackref,breaklinks,colorlinks]{hyperref}

\usepackage[capitalize]{cleveref}
\crefname{section}{Sec.}{Secs.}
\Crefname{section}{Section}{Sections}
\Crefname{table}{Table}{Tables}
\crefname{table}{Tab.}{Tabs.}

\begin{document}

\title{OcclusionFusion: Occlusion-aware Motion Estimation\\ for Real-time Dynamic 3D Reconstruction}

\author{
    Wenbin~Lin$\qquad\qquad$
    Chengwei~Zheng$\qquad\qquad$
    Jun-Hai~Yong\textsuperscript{*}$\qquad\qquad$
    Feng~Xu\textsuperscript{\thanks{This work was supported by Beijing Natural Science Foundation (JQ19015), the NSFC (No.61727808, 62021002), the National Key R\&D Program of China (2018YFA0704000, 2019YFB1405703) and TC190A4DA/3. This work was supported by THUIBCS, Tsinghua University and BLBCI, Beijing Municipal Education Commission. Jun-Hai Yong and Feng Xu are corresponding authors.}} \\
    School of Software and BNRist, Tsinghua University \\
    {\tt\small lwb20@mails.tisnghua.edu.cn, \{zhengcw18, xufeng2003\}@gmail.com, yongjh@tsinghua.edu.cn}
}

\maketitle

\input{sec/0-abstract}
\input{sec/1-introduction}
\input{sec/2-related-work}
\input{sec/3-method}
\input{sec/4-experiments}
\input{sec/5-limitations}
\input{sec/6-conclusion}

{\small
\bibliographystyle{ieee_fullname}
\bibliography{main}
}

\newpage
\input{sec/7-appendix}

\end{document}

%% file: sec/0-abstract.tex
\begin{abstract}
RGBD-based real-time dynamic 3D reconstruction suffers from inaccurate inter-frame motion estimation as errors may accumulate with online tracking. This problem is even more severe for single-view-based systems due to strong occlusions. Based on these observations, we propose OcclusionFusion, a novel method to calculate occlusion-aware 3D motion to guide the reconstruction. In our technique, the motion of visible regions is first estimated and combined with temporal information to infer the motion of the occluded regions through an LSTM-involved graph neural network. Furthermore, our method computes the confidence of the estimated motion by modeling the network output with a probabilistic model, which alleviates untrustworthy motions and enables robust tracking. Experimental results on public datasets and our own recorded data show that our technique outperforms existing single-view-based real-time methods by a large margin. With the reduction of the motion errors, the proposed technique can handle long and challenging motion sequences. Please check out the project page for sequence results: \url{https://wenbin-lin.github.io/OcclusionFusion}.
\end{abstract}

%% file: sec/1-introduction.tex
\section{Introduction}
Dynamic 3D reconstruction has been attracting more and more attention with the development of sensing and computing techniques. It aims to reconstruct the shape, motion, and appearance of the recorded objects, and thus enables users to record, edit, animate, and play with real 3D targets for various applications, including 3D design, video games, telecommunications, virtual reality, and augmented reality. 

In dynamic 3D reconstruction with RGB-D sensors, fusion-based works ~\cite{newcombe2015dynamicfusion, volumedeform, guo2017real, deepdeform, nnrt} have achieved impressive results in recent years and have become a new technological trend with several important features.
Firstly, these techniques do not require geometry templates of the target objects but fuse the geometries online with the recording. 
Therefore, they can handle various targets, including humans, objects, and even 3D scenes~\cite{bundlefusion, mixfusion}. 
Secondly, they can handle both rigid and nonrigid motions without requiring class-specific motion priors. 
This is also important to increase the generalization capability of 3D reconstruction. 
Thirdly, they can be achieved in real time and with a single consumer sensor, which makes these techniques be easily used by end-users.

\begin{figure}[]
	\centerline{\includegraphics[width=1.0\linewidth]{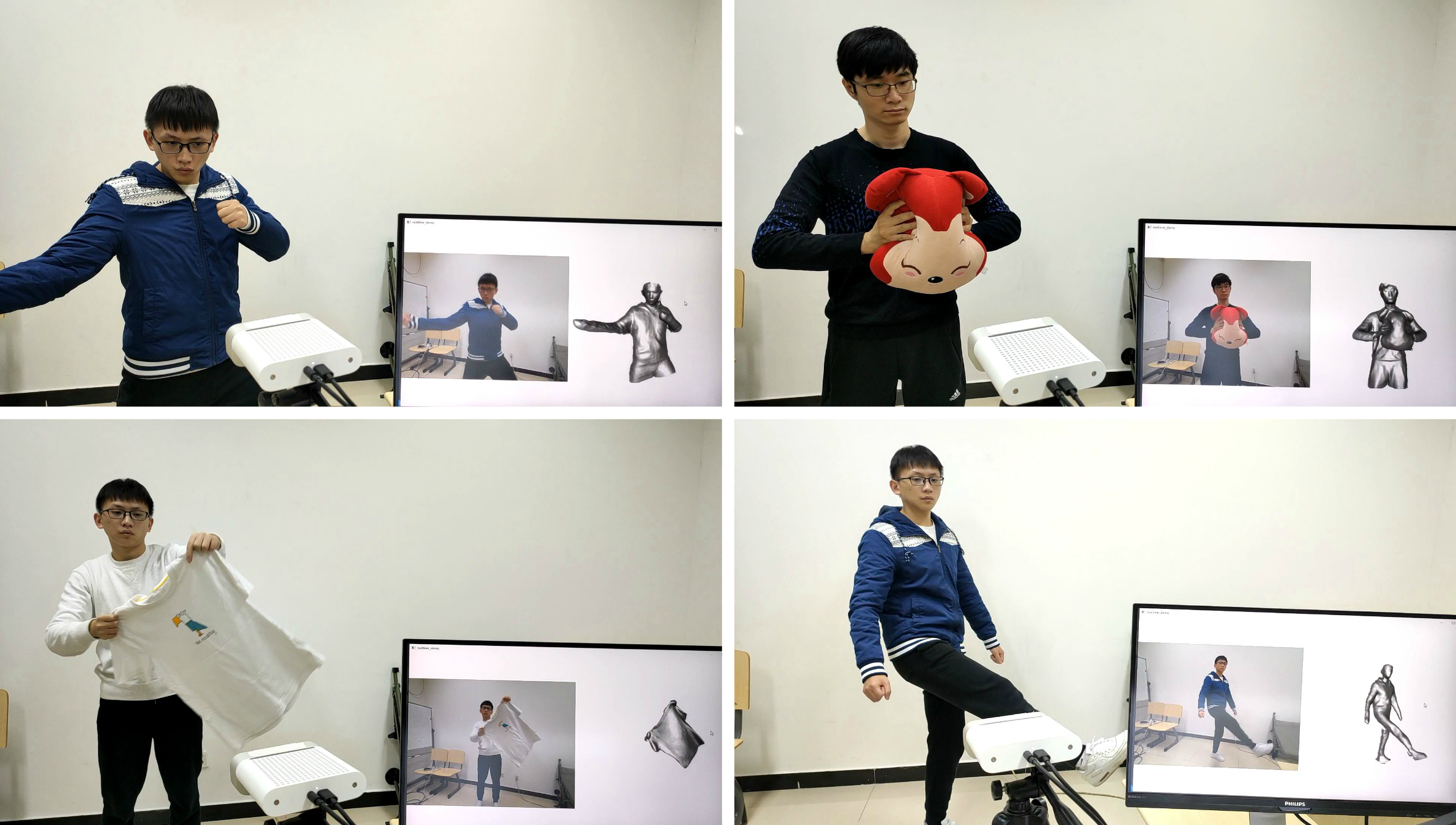}}
	\caption{Our method can reconstruct dynamic objects in real time.}
	\label{fig:teaser}
\end{figure}

In single view-based solutions, there is a significant quality gap between online and offline methods. 
On the online side, existing methods use iterative geometry fitting~\cite{newcombe2015dynamicfusion}, sparse image feature matching~\cite{volumedeform}, or photometric constrains~\cite{guo2017real} to estimate object motions. 
However, these real-time techniques cannot give very reliable temporal correspondences, and thus the accuracy of motion estimation is limited. 
With the error accumulation, these techniques tend to fail to track long sequences or challenging motions.
On the other hand, offline methods can build much more accurate temporal correspondences without considering the computation complexity~\cite{deepdeform, nnrt}, which leads to much better reconstruction results. 

We propose OcclusionFusion, which fills the quality gap between online and offline methods with real-time performance (Fig.~\ref{fig:teaser}) and the state-of-the-art accuracy even compared to offline methods. 
This is accomplished by better and efficiently exploring the spatial and temporal motion priors. 
For single view-based 3D dynamic reconstruction, occlusion is one key obstacle as the occluded regions need to be reconstructed without motion observations. 
On the other hand, we know that the occluded regions do not move arbitrarily. The motion of the visible regions and the historical motion information give strong prior knowledge to constrain the motion estimation of the occluded regions. 
Based on this observation, we propose to train a neural network to estimate full 3D motions of whole objects including the occluded surfaces using the motions of the visible regions as well as the historical information. 
With the obtained full object motion between consecutive frames, we do not require either exhaust correspondence computation like DeepDeform~\cite{deepdeform} or long range correspondences between multiple frames like Bozic et al.~\cite{nnrt}. Either of them involves heavy computation costs that hinder real-time performance.

To estimate the motion of the occluded regions, we propose to train a light-weight graph neural network.
The graph neural network integrates both the motion of the visible region by the graph structure and the historical information by involving a long short-term memory (LSTM)~\cite{lstm} module.
Recent work 4DComplete~\cite{4dcomplete} predicted the motion beyond the observable regions by a 3D convolution-based neural network.
However, the 3D convolution module requires high computation and memory costs, which prevents their method from achieving real-time performance.
 
Furthermore, we model the per graph node motion using a Gaussian distribution, which not only improves the accuracy of motion prediction but also provides a confidence to aid the reconstruction module. 
With the confidence, we down-weight untrustworthy motion and improve the robustness of the reconstruction system.

In summary, the contributions lie in three aspects:
\setlist{nolistsep}
\begin{itemize}[noitemsep]
    \itemsep0em
    \item 
        We proposed a robust real-time dynamic 3D reconstruction system with a light-weight graph neural network for full 3D motion estimation. Various results including the one on public benchmark show that our real-time system outperforms the state-of-the-art offline methods.
    \item 
    The graph neural network involves LSTM structure to leverage both the spatial and temporal information to predict the full object motion accurately and efficiently.
    \item 
    Per node motion confidence is estimated by modeling the predicted motion using a Gaussian distribution, which gives more information for the reconstruction system to achieve high robustness.
\end{itemize}

\begin{figure*}[t]
\begin{center}
   \includegraphics[width=1.0\linewidth]{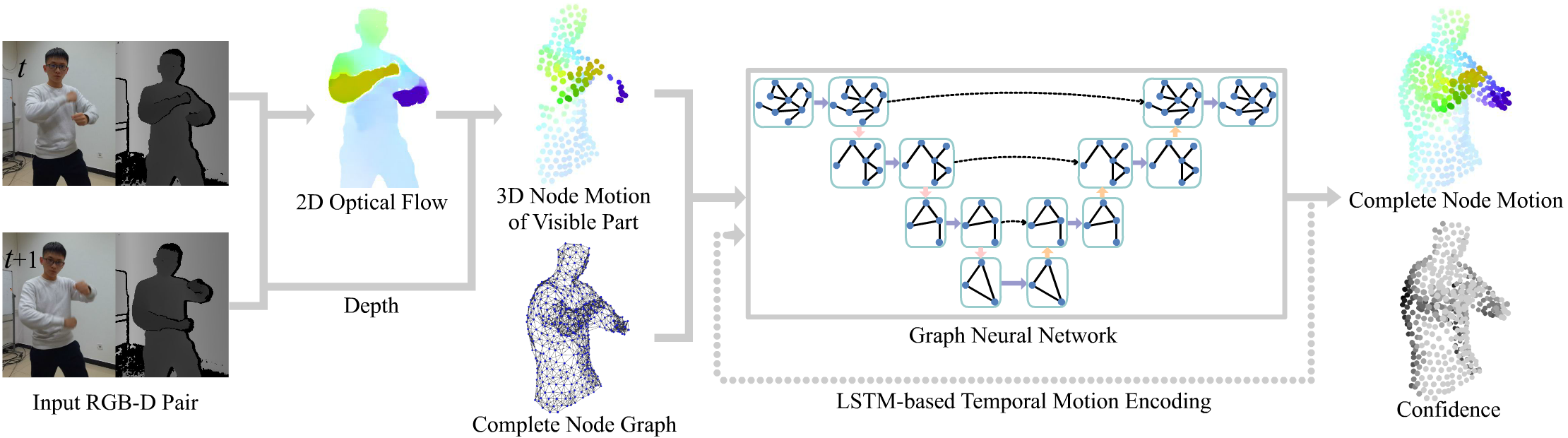}
\end{center}
\caption{Pipeline of complete node motion estimation. 
Given input RGB-D images at frame $t$ and $t+1$ as well as the complete node graph at frame $t$, we first use a neural network to estimate the 2D optical flow between the input image pair.
Then, we combine the 2D optical flow and the depth images to compute the 3D motion of visible nodes. 
The graph neural network uses both the motion of the visible nodes and the complete node graph as input.
Meanwhile, the graph neural network integrates the historical node motion by involving an LSTM module to estimate complete node motion and per node confidence.
The color encoding of optical flow is from ~\cite{flowcolor}.}
\label{fig:achitecture}
\end{figure*}

%% file: sec/2-related-work.tex
\section{Related Work}
\subsection{RGB-D Based Dynamic Reconstruction}
Reconstruction of non-rigid deforming objects using a single RGB-D camera is an important research area in computer vision and graphics. 
Free-form capture~\cite{newcombe2015dynamicfusion, volumedeform, guo2017real, killingfusion, SobolevFusion, deepdeform, nnrt} does not assume any geometric prior and can reconstruct general non-rigid scenes.

DynamicFusion~\cite{newcombe2015dynamicfusion} is the pioneering work that ables to capture non-rigid scenes in real-time by a hierarchical node graph structure and an efficient GPU solver. However, DynamicFusion did not utilize any visual cues from color images and cannot track the target robustly. 
VolumeDeform~\cite{volumedeform} introduced a grid structure to represent the object deformation and used sparse SIFT~\cite{sift} features to improve the registration quality. 
Guo et al.~\cite{guo2017real} proposed a joint geometry, albedo, and motion field optimization pipeline to obtain a higher quality. 
KillingFusion~\cite{killingfusion} and SobolevFusion~\cite{SobolevFusion} are able to handle topology changes but cannot recover space-time correspondences along the whole input sequence. 
DeepDeform~\cite{deepdeform} and Bozic et al.~\cite{nnrt} leveraged learning-based correspondences to improve the ability to handle fast motions.
However, the exhaust correspondences prediction process of DeepDeform, and the multi-frame matching strategy of Bozic et al.~\cite{nnrt} are time-consuming and hinder the reconstruction system from real-time.
Other methods use a multi-camera setup like Fusion4D~\cite{fusion4D} or high frame-rate depth cameras like Motion2Fusion~\cite{motion2fusion} to improve the reconstruction results. 
However, the equipment is difficult to access by average consumers.
We present a novel real-time dynamic reconstruction method that can handle challenging motion with occlusion using only a single consumer depth sensor.

\subsection{Learning-based Motion Estimation}
In recent years, learning-based motion estimation methods have shown great superiority compared with hand-craft image features like SIFT~\cite{sift} and SURF~\cite{surf} as well as hand-crafted geometry descriptors~\cite{Andrea2004, Andrew1999, Federico2010, Radu2008, Radu2009}.
Optical flow~\cite{dosovitskiy2015flownet, flownet2, selflow, pwcnet, raft} and scene flow~\cite{pointflow, deeprigidflow, flownet3d, flownet3dpp} methods achieve promising results in dense motion estimation. 
DeepDeform~\cite{deepdeform} trained a delicate network to estimate sparse temporal correspondences for pixels by using the corresponding patch information. 
Bozic et al.~\cite{nnrt} imposed dense optical flow to the reconstruction system, which is much more efficient than DeepDeform. 
However, the methods mentioned above only predict the motion of the visible part. 
Recently, 4DComplete~\cite{4dcomplete} created a synthetic animation dataset called DeformingThings4D and proposed the first approach that predicts motion beyond the observable surface. However, its 3D convolution module is computationally intensive and cannot be applied to real-time scenarios like our method.

\subsection{Graph Neural Networks}
Graph Neural Networks (GNNs) can model non-Euclidean data structure and have been applied to various scenarios like social networks~\cite{gnn_social}, protein-protein interaction prediction~\cite{protein}, 3D pointcloud analysis~\cite{edgeconv}, etc. 
Among existing works, graph convolutional networks (GCN)~\cite{gcn} performed spectral convolutions on graphs to propagate information among nodes. 
Graph U-Net~\cite{graph-u-net} proposed graph pooling and unpooling operations based on trainable similarity measurements. 
Li et al.~\cite{deepgcn} leveraged residual connections to train deeper GNN and solve the problem of vanishing gradient and over-smoothing. 
Shi et al.~\cite{transformer_conv} adopted multi-head attention to the message passing of graph learning.
We adopt the skip connection in ~\cite{deepgcn} and graph transformer module in ~\cite{transformer_conv} in our network. 
Similar to Graph U-Net, our method constructs a graph pyramid, while the feature upsampling and downsampling process of our approach is precomputed and untrainable.

%% file: sec/3-method.tex
\section{Method}
We build a single-view RGB-D based dynamic 3D reconstruction system which reconstructs a volumetric model and solves non-rigid surface motion frame-by-frame. 
The output motion is parameterized by the deformation of a node graph. 
To handle the motion tracking of the occluded nodes, we design an occlusion-aware motion estimation network (Sec.~\ref{sec:network}). 
The network infers the motion and the motion confidence of the whole node graph based on spatial and temporal motion observation. 
The pipeline of complete node motion estimation is demonstrated in Fig.~\ref{fig:achitecture}.
Given an input RGB-D image and the node graph of the current reconstructed model, we first estimate the 2D optical flow from the previous image to the current input image with a neural network.
Then, with the 2D optical flow and the depth images, we compute the 3D motion of the visible nodes. 
Both the 3D motion of the visible nodes and the complete node graph are fed to the graph neural network.
Meanwhile, we involve an LSTM module to integrate the historical motion of the nodes, which carries temporal information to the graph neural network.
Finally, the graph neural network predicts full node motion with per node confidence.
With the predicted node motion and the confidence, we further optimize the deformation parameters of graph nodes and build a robust non-rigid reconstruction system (Sec.~\ref{sec:reconstruction}).

\subsection{Occlusion-aware Motion Estimation Network}
\label{sec:network}
We use the graph-based representation proposed in ~\cite{SumnerSP07} to parameterize object motion with sparse graph nodes. 
The graph nodes are uniformly sampled on the object surface.
Points on the object surface are driven by the deformation of the graph nodes.
However, with a single-view RGB-D camera, the motion of graph nodes is hard to solve when occlusions occur. 
To solve this problem, we propose an occlusion-aware motion estimator using a graph neural network.

\noindent\textbf{Network Architecture.} 
The graph neural network conducts message passing among nodes based on the node connectivity of the input node graph. 
As shown in Fig.~\ref{fig:achitecture}, the goal of the graph neural network is to predict the motion of the whole node graph with per node confidence.
The network's input contains the complete graph structure, the motion of the visible nodes, and the motion of the historical frames. 

Specifically, the input node feature includes the node's 3D position and the 3D motion if the node is visible. 
For the occluded nodes, we assign their motion as zeros. 
Besides, we add an extra dimension to indicate the node's visibility. The value will be 1 for visible and 0 for occluded. 
In addition, we extract the rigid motion from the total motion by computing a rigid transformation in $\mathrm{SE}(3)$ based on the motion of the visible nodes.
Therefore, the neural network only needs to predict the non-rigid motion.

As the observation of a single frame is not strong enough to constrain the occluded motion, we combine the historical motion and the observation of the current frame to compute the features of the graph nodes.
To integrate historical motion, we maintain a motion state for each node by involving an LSTM module. 
For each node, the LSTM uses the estimated historical motion and confidence to predict the node motion and confidence at the current frame.

To model 3D node motion with confidence, we use a Gaussian distribution with diagonal covariance $\mathcal{N}(\boldsymbol{\mu}, \sigma^2\boldsymbol{I})$ to represent the motion of each node, where $\boldsymbol{\mu}$ is the predicted 3D motion, $\sigma$ is the standard deviation used to reflect the confidence. 

For message passing among nodes, we construct a multi-scale graph pyramid $\{ \mathcal{G}_1, \mathcal{G}_2, \mathcal{G}_3, \mathcal{G}_4\}$ based on the input node graph.
The network contains node feature downsampling, upsampling and skip connection between the same level of the pyramid like Graph U-Net~\cite{graph-u-net}.
The basic graph convolution module is the graph transformer proposed by Shi et al.~\cite{transformer_conv}. 
Besides, we add a residual connection to the graph convolution module like Li et al.~\cite{deepgcn}. 
More details of the network can be found in the supplemental document.

\noindent\textbf{Network Training.} 
To supervise the network training, we model the target node motion with the predicted Gaussian distribution and use a standard log-likelihood loss:
\begin{equation}
\begin{aligned}
    \mathcal{L}_\mathrm{out}(\boldsymbol{\Theta}_\mathrm{lstm}, \boldsymbol{\Theta}_\mathrm{graph}) = -\frac{1}{N}\sum_{i=1}^N \log \left(\mathcal{N}\left(\boldsymbol{y}_i \mid \boldsymbol{\mu}_{i}, \sigma_{i}^2 \boldsymbol{I}\right)\right), 
\end{aligned}
\end{equation}
where $\boldsymbol{\Theta}_\mathrm{lstm}$ and $\boldsymbol{\Theta}_\mathrm{graph}$ are the parameters of the LSTM module and the graph neural network, $\boldsymbol{\mu}_i$ and $\sigma_i$ are the final outputs of the network, $N$ is the number of nodes in the node graph, $\boldsymbol{y}_i$ is the ground truth motion of the $i$th node. 

For temporal motion encoding of the LSTM module, we impose another log-likelihood loss to the outputs of the LSTM module:
\begin{equation}
    \mathcal{L}_\mathrm{temp}(\boldsymbol{\Theta}_\mathrm{lstm}) = -\frac{1}{N}\sum_{i=1}^N \log \left(\mathcal{N}\left(\boldsymbol{y}_i \mid \boldsymbol{\mu}'_{i}, {\sigma'_{i}}^2 \boldsymbol{I}\right)\right),
\end{equation}
where $\boldsymbol{\mu}'_i$ and $\sigma'_i$ are the outputs of the LSTM module. 
Besides, we find that truncating the minimum value of $\sigma_i$ and $\sigma'_i$ to 0.1 leads to slightly better performance.

The total loss is a weighted combination of the above two terms:
\begin{equation}
    \mathcal{L_\mathrm{total}} = \mathcal{L}_\mathrm{out} + 0.1 \mathcal{L}_\mathrm{temp}.  
\end{equation}

\subsection{Confidence Guided Non-Rigid Reconstruction}
\label{sec:reconstruction}
For non-rigid RGB-D reconstruction, we follow the classic volumetric pipeline that uses a truncated signed distance field (TSDF) to store the canonical model and a motion field to warp the canonical model to align it with the input image sequence.

We use the graph-based representation proposed in ~\cite{SumnerSP07} to parameterize the non-rigid motion. 
The motion field $\mathcal{W}$ can be and represented by a deformation graph $\mathcal{G} = \{[\boldsymbol{p}_i, \boldsymbol{T}_i]\}$, where $\boldsymbol{p}_i\in \mathbb{R}^3$ is the position of the $i$th node, $\boldsymbol{T}_i \in \mathrm{SE}(3)$ is the transformation of the $i$th node. 
The motion field can be computed by convex combinations of neighboring nodes' transformations.

Given the reconstructed model $\mathcal{M}^{t-1}$ and the input RGB-D image pair $\{\mathcal{C}^{t-1}, \mathcal{D}^{t-1}\}$ and $\{\mathcal{C}^{t}, \mathcal{D}^{t}\}$, we optimize the following energy to solve $\mathcal{W}^{t}$.
\begin{equation}
\label{eq:energy_total}
\begin{aligned}
     E_\mathrm{total}(\mathcal{W}^t) = & \lambda_\mathrm{depth}E_\mathrm{depth} + \lambda_\mathrm{motion}E_\mathrm{motion} + \\ & 
     \lambda_\mathrm{2D}E_\mathrm{2D} + \lambda_\mathrm{reg}E_\mathrm{reg}.
\end{aligned}
\end{equation}

For the depth term $E_\mathrm{depth}$, we employ dense point-to-plane alignment between depth image $\mathcal{D}^{t}$ and the warped model:
\begin{equation}
    E_\mathrm{depth} = \sum_{(\boldsymbol{v}, \boldsymbol{u}^t) \in \mathcal{P}} (\boldsymbol{n}_{\boldsymbol{u}^t}^\text{T} (\boldsymbol{v}'-\boldsymbol{u}^t))^2, 
\end{equation}
where $\boldsymbol{v}$ is a vertex on canonical model $\mathcal{M}$, $\boldsymbol{v}'$ is the warped vertex given by $\mathcal{W}^t$. 
$\boldsymbol{u}^t$ is a 3D point projected from the depth image $\mathcal{D}^t$, its normal is represented as $\boldsymbol{n}_{\boldsymbol{u}^t}$.
$\mathcal{P}$ is the vertex pair set. 
We first render the warped model $\mathcal{M}^{t-1}$ to camera view at $t-1$ and get the projected 2D coordinate of all visible vertices. 
Let $\Pi$ be the projection function, for $\Pi(\boldsymbol{v}')$, we use the computed 2D optical flow to find the corresponding pixel where $\Pi(\boldsymbol{u}^t)$ locates.

Based on the output motion and confidence of the neural network, we constrain the motion field to be close to the network's prediction and construct a motion energy term as follow:
\begin{equation}
    E_\mathrm{motion} = \sum_{i \in \mathcal{G}} w_i \left\|\boldsymbol{T}_i^t \boldsymbol{p}_i - \left( \boldsymbol{T}_i^{t - 1} \boldsymbol{p}_i + \boldsymbol{\mu}_i \right) \right\|^2_2, 
\end{equation}
where $\boldsymbol{\mu}_i$ denotes the motion predicted by the neural network at the $i$th node, $w_i$ is the weight computed through $\sigma_i$ and $\boldsymbol{\mu}_i$:
\begin{equation}
\label{eq:energy_weight}
    w_i = \exp \left(-\frac{k\sigma_i^2}{\left(\left\| \boldsymbol{\mu}_i \right\|_2 + \epsilon \right)^2} \right).
\end{equation}

Given the dense correspondences between consecutive images, the 2D term constrains the projection of warped vertices to be consistent with the correspondences of the 2D optical flow:
\begin{equation}
    E_\mathrm{2D} = \sum_{(\boldsymbol{v}, \boldsymbol{u}^t) \in \mathcal{P}} \|\Pi(\boldsymbol{v}')-\Pi(\boldsymbol{u}^t)\|^2_2.
\end{equation}

$E_\mathrm{reg}$ is the as-rigid-as-possible regularizer, 
\begin{equation}
    E_\mathrm{reg} = \sum_{j\in \mathcal{G}}\sum_{i \in N_j} \| \boldsymbol{T}_j^t\boldsymbol{p}_i - \boldsymbol{T}_i^t\boldsymbol{p}_j \|^2_2, 
\end{equation}
where $N_j$ denotes the neighboring nodes of the $j$th node.

After solving the motion field $\mathcal{W}^t$, we update the canonical TSDF volume.
If there is a region beyond the coverage of the current node graph, new nodes will be inserted. 
For details on geometry fusion and node graph extension, please refer to DynamicFusion~\cite{newcombe2015dynamicfusion}.


\subsection{Implementation Details}
\label{sec:details}
\noindent\textbf{Multi-scale Graph Pyramid Construction.} 
Given a node graph sampled on the object surface, we construct a multi-scale graph pyramid by subsampling the nodes with different node intervals.
The nodes in the $(l+1)$th level are a subset of the nodes in the $l$th level. 
The higher the level, the larger the interval. 
As for the edges between nodes, we compute $k$-nearest neighbors of each node in the Euclidean space at the first level. 
Then we use the breadth-first search on the graph to compute neighbors for higher levels. 
More details can be found in the supplemental document.

\noindent\textbf{Data Generation.} 
Network training relies on the supervision of the full 3D motion of objects. 
We leverage the DeformingThings4D~\cite{4dcomplete} dataset, which contains 1,972 animations spanning 31 categories of humanoids and animals.

To simulate the 3D reconstruction scenario, we first uniformly sample graph nodes on the object's surface. 
Then, we introduce a virtual camera that keeps the moving object in the center of the rendered depth image. 
Once a graph node is visible to the camera, the node will be added to an observed node set $\mathcal{N}$. 
Therefore, the number of the observed nodes will increase over time. For frame $t$ we build a graph pyramid of the current observed node set $\mathcal{N}^t$. 
Based on the nodes' visibility, $\mathcal{N}^t$ can be divided into $\mathcal{N}^t_\text{vis}$ and $\mathcal{N}^t_\text{occ}$. 
For each node, we compute the 3D displacement between temporally adjacent frames as its motion, and only the motion of $\mathcal{N}^t_\text{vis}$ is fed to the network. 
In addition, we randomly resize the objects into a 3D bounding box ranging from $1\mathrm{m}\times 1\mathrm{m} \times 1\mathrm{m}$ to $2\mathrm{m}\times 2\mathrm{m} \times 2\mathrm{m}$, which is a common volume for dynamic reconstruction systems.

\noindent\textbf{3D Reconstruction System.} 
In the 3D reconstruction system, we leverage state-of-the-art 2D optical flow prediction method RAFT~\cite{raft} to estimate the optical flow between the input image pairs. 
As the input images contain depth information, we retrain the RAFT network with RGB-D four-channel input on synthetic optical flow dataset Sintel~\cite{sintel}, FlyingThings3D\cite{flyingthings}, and Monkaa~\cite{flyingthings}. 
The extra depth channel helps to achieve higher accuracy, especially when motion blur occurs in the color images. 
The choice of 2D optical flow methods is discussed in the supplemental document.
Then, with the help of depth images, we compute the visibility of graph nodes and backproject the 2D image coordinate to 3D space to obtain the 3D motion of the visible nodes. 
For the 3D motion of the historical frames, we use the output of the 3D reconstruction system rather than the network's output, as it integrates more constraints and is usually more accurate.

\begin{table}
    \begin{center}
    \begin{tabular}{lcc}
        \hline
        Method & EPE(mm) \\
        \hline
        Rigid Fitting & 15.47 \\
        ARAP Deformation~\cite{SumnerSP07} & 4.39 \\
        \hline
        Ours w/o temporal motion & 4.02 \\
        Ours w/o log-likelihood loss & 4.07 \\
        Ours & 3.75 \\
        Ours + post-processing & \textbf{3.28} \\
        \hline
    \end{tabular}
    \end{center}
    \caption{Motion estimation results of the occluded nodes.}
	\label{tab:dt4d}
\end{table}

\begin{table}
    \begin{center}
    \resizebox{1.0\linewidth}{!}
    {
        \begin{tabular}{llcc}
        \hline
        & Method & Geo. error & Def. error \\ 
        \hline
        \multirow{4}{*}{Online}
        & DynamicFusion~\cite{newcombe2015dynamicfusion} & 1.078 & 6.179 \\
        & VolumeDeform~\cite{volumedeform}  & 7.485 & 20.841 \\
        & Ours w/o motion term & 0.406 & 3.154 \\
        & Ours & \textbf{0.386} & \textbf{2.800} \\ 
        \hline
        \multirow{2}{*}{Offline} 
        & DeepDeform~\cite{deepdeform} & 0.416 & 3.152 \\
        & Bozic et al.~\cite{nnrt} & 0.403 & 2.872 \\ 
        \hline
        \end{tabular}
    }
    \end{center}
    \caption{Results on the DeepDeform~\cite{deepdeform} non-rigid reconstruction benchmark. The geometry error evaluates the difference between the reconstructed model and the input depth image. The deformation error evaluates the consistency between the motion tracking results and manually annotated correspondences. All values are in cm.}
	\label{tab:deepdeform}
\end{table}

\begin{figure}[]
	\centerline{\includegraphics[width=0.90\linewidth]{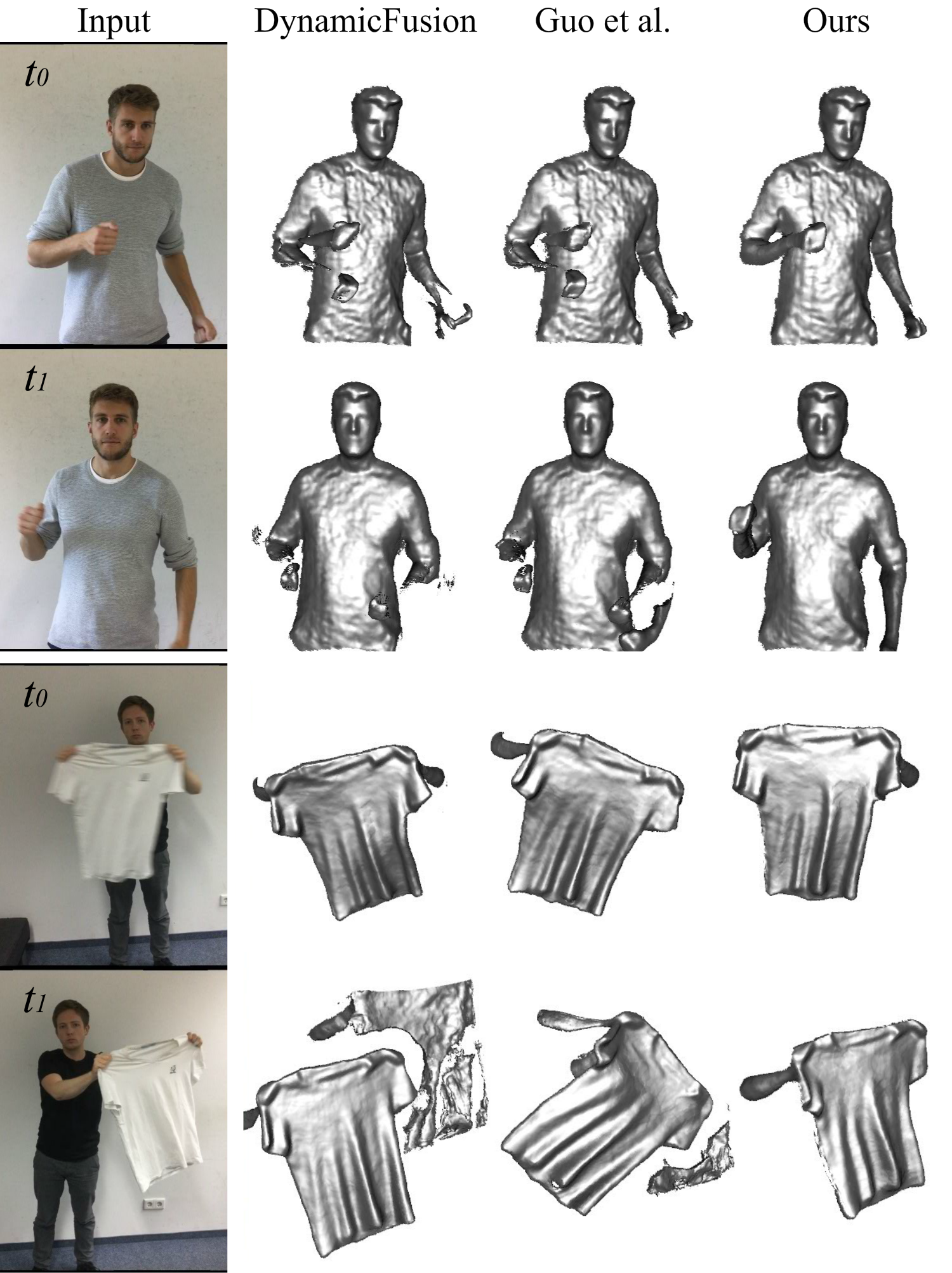}}
	\caption{Qualitative comparisons with DynamicFusion~\cite{newcombe2015dynamicfusion} and Guo et al.~\cite{guo2017real} on the DeepDeform dataset.}
	\label{fig:cmp_online}
\end{figure}

\begin{figure}[]
	\centerline{\includegraphics[width=0.90\linewidth]{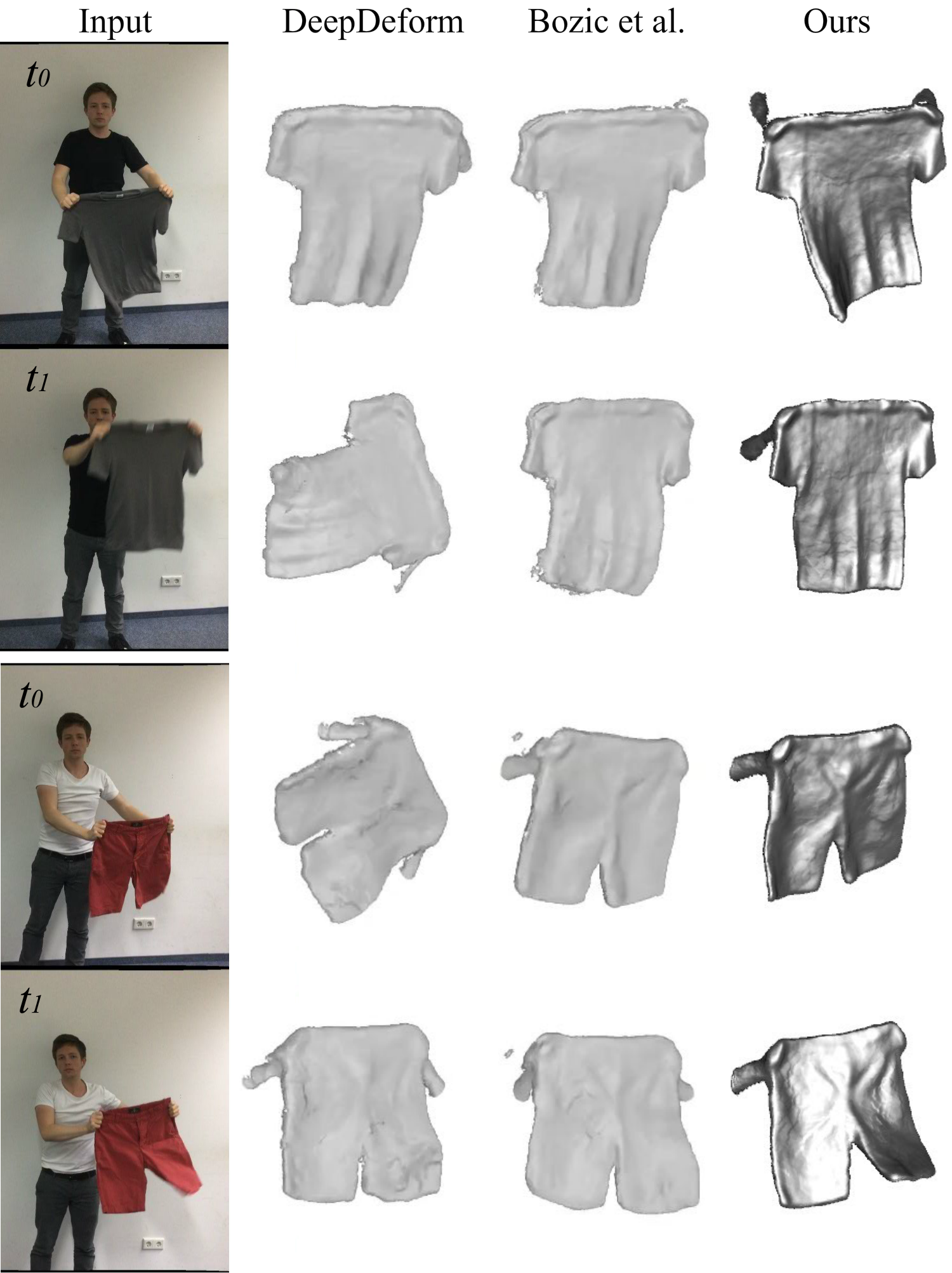}}
	\caption{Qualitative comparisons with DeepDeform~\cite{deepdeform} and Bozic et al.~\cite{nnrt} on the DeepDeform dataset. Results of DeepDeform and Bozic et al.~\cite{nnrt} come from the video of Bozic et al.~\cite{nnrt}.}
	\label{fig:cmp_offline}
\end{figure}

%% file: sec/4-experiments.tex
\section{Experiments}
In this section, we first present the performance and parameters of our system (Sec.~\ref{sec:performance}).
Then, we evaluate the accuracy of the occlusion-aware motion prediction network (Sec.~\ref{sec:motion_eval}).
Next, we evaluate the dynamic reconstruction system quantitatively and qualitatively (Sec.~\ref{sec:rec_eval}).
Finally, in our accompanying video, we show sequence results to fully demonstrate the power of our technique.

\subsection{Performance and Parameters}
\label{sec:performance}
Our system is implemented on a computer with an Intel i7-10700 CPU, 32 GB RAM, and two NVIDIA RTX GeForce 2080Ti graphics cards. 
We use an Azure Kinect DK to record RGB-D sequences at $30$ fps. 
The depth and color images are aligned and resized to $640\times 480$. 
The full pipeline runs in real time at about $36\mathrm{ms}$ per frame, where the neural network-based motion prediction takes $13\mathrm{ms}$, warp field solving takes $16\mathrm{ms}$, and $7\mathrm{ms}$ for model update, marching cubes, and anything else. 
The 2D optical flow estimation step takes about $26\mathrm{ms}$ on another graphics card. 
Since we have made the two steps in a pipeline, our system achieves real-time performance but with a one-frame delay.

The weights in Eq.~\ref{eq:energy_total} are empirically set as $\lambda_\mathrm{depth}=1, \lambda_\mathrm{motion}=2, \lambda_\mathrm{2D}={1e-6}, \lambda_\mathrm{reg}=5$. 
$k$ and $\epsilon$ in Eq.~\ref{eq:energy_weight} are set to $4$ and $1\mathrm{cm}$.

\subsection{Experiments of Motion Prediction}
\label{sec:motion_eval}
We evaluate the accuracy of motion prediction on the  DeformingThings4D~\cite{4dcomplete} dataset. 
Given a node set $\mathcal{N}$, its subset of visible nodes of a given camera viewpoint, and the 3D motion of the visible nodes, the goal is to estimate the motion of the occluded nodes. 
Following 4DComplete~\cite{4dcomplete}, we use rigid fitting and as-rigid-as-possible (ARAP) deformation as baselines to compare with ours. 
The rigid fitting approach estimates the occluded motion by a global rigid transformation estimated by the visible motion.
The ARAP deformation approach leverages local rigidity and the motion of visible nodes to optimize the occluded motion.
Here we use a graph-based ARAP deformation proposed by~\cite{SumnerSP07}. 
For evaluation, we randomly choose 100 animations from the animal motion subset, which are never used during training, and we apply the average 3D end-point-error (EPE) over all 100 sequences as the metric.

\noindent\textbf{Comparisons.} 
The results are reported in Tab.~\ref{tab:dt4d}. 
Rigid fitting gives the worst result because there are large non-rigid motions in the test sequences.
ARAP deformation cannot give good results either because it only considers local rigidity priors.
On the other hand, we achieve the best performance because we explore multi-scale motion priors through the graph pyramid structure.
Like 4DComplete, we find that employing optimization-based post-processing with ARAP prior further improves the results. 

\noindent\textbf{Evaluations.} 
Furthermore, we train the network without the temporal motion integrated by the LSTM, and the error increases, which verifies the effectiveness of the temporal motion.
In addition, we use mean-squared error (MSE) loss instead of log-likelihood loss based on per node Gaussian distribution to train the network, and the error is also larger.
In this case, the MSE loss is equivalent to the log-likelihood loss with a fixed $\sigma$ for every node.
As the motion uncertainties of the visible nodes and the occluded nodes are obviously different, we believe using per node Gaussian distribution to model the node motion is more appropriate. 
Note that the network is trained only with humanoid motions. It still performs well on animal sequences, which shows the strong generalization ability of the network.

Since the data preprocessing and network training of 4DComplete~\cite{4dcomplete} are not yet publicly available, we are not able to make a direct comparison with their method.
According to the results in Tab.~\ref{tab:dt4d}, our approach outperforms ARAP deformation by a large margin, which is similar to 4DComplete.
However, our light-weight neural network takes only $13\mathrm{ms}$ for motion and confidence prediction, while 4DComplete takes approximate $3\mathrm{s}$ and cannot predict confidence, which is important guidance for 3D dynamic reconstruction as indicated in our experiments (Sec~.\ref{sec:ablation}).

\begin{figure}[]
	\centerline{\includegraphics[width=0.89\linewidth]{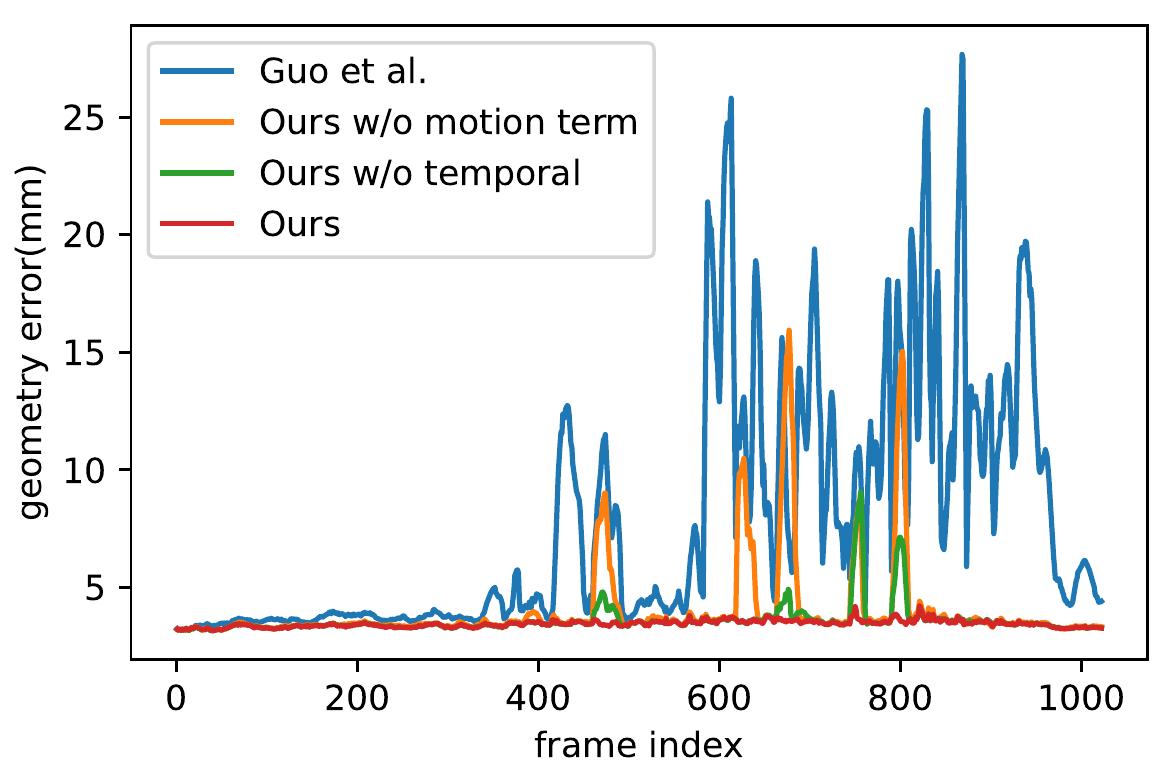}}
	\caption{Geometry errors on the top sequence of Fig.~\ref{fig:ablation_seq}. The average geometry errors over the whole sequence are $7.68\mathrm{mm}$, $4.01\mathrm{mm}$, $3.58\mathrm{mm}$, and $3.45\mathrm{mm}$ from Guo et al.\cite{guo2017real} to Ours.}
	\label{fig:geo_err}
\end{figure}

\begin{figure}[]
	\centerline{\includegraphics[width=0.88\linewidth]{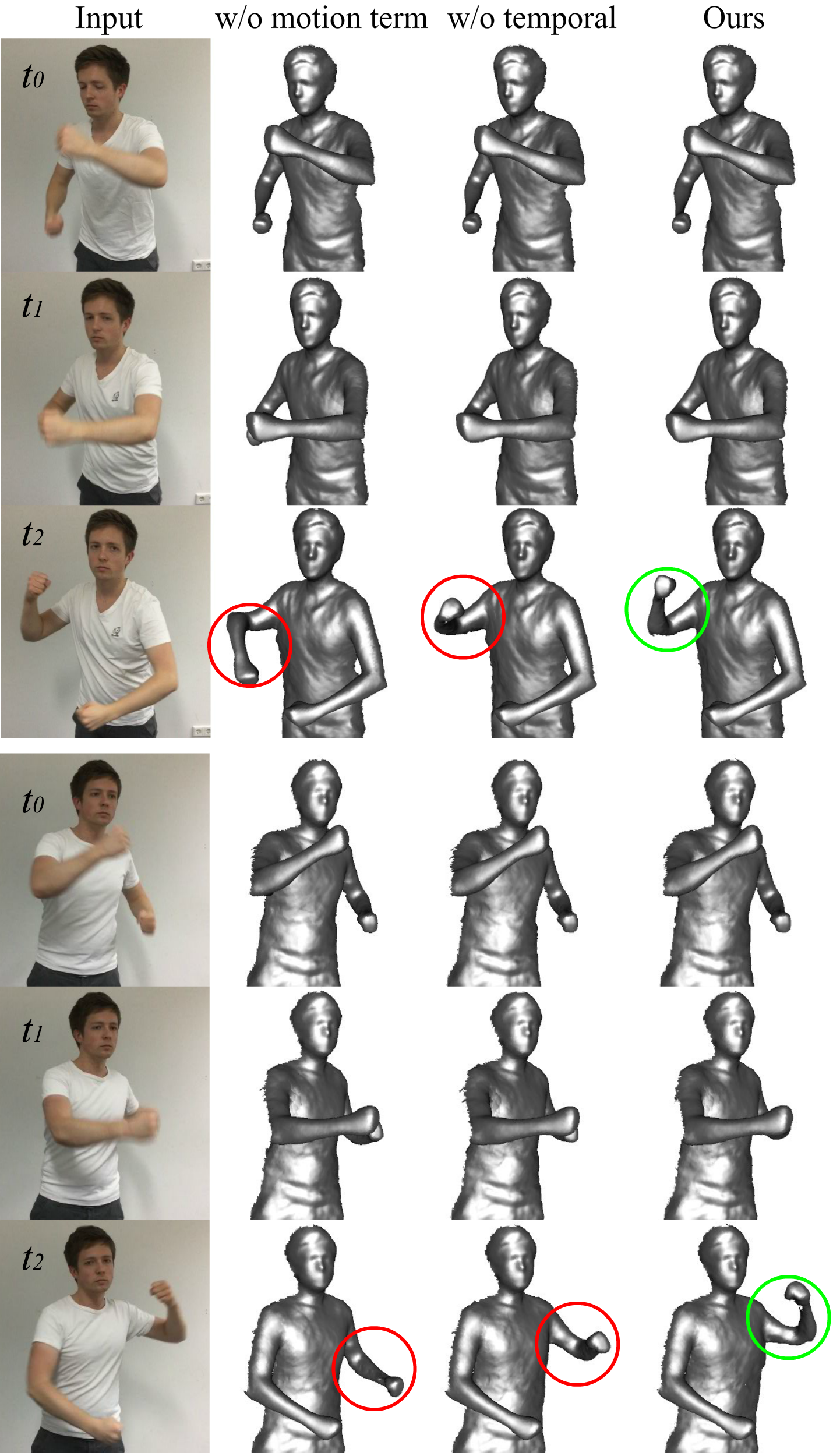}}
	\caption{Ablation study on $E_\mathrm{motion}$ and temporal information. Both sequences are from the DeepDeform dataset.}
	\label{fig:ablation_seq}
\end{figure}

\begin{figure}[]
	\centerline{\includegraphics[width=0.90\linewidth]{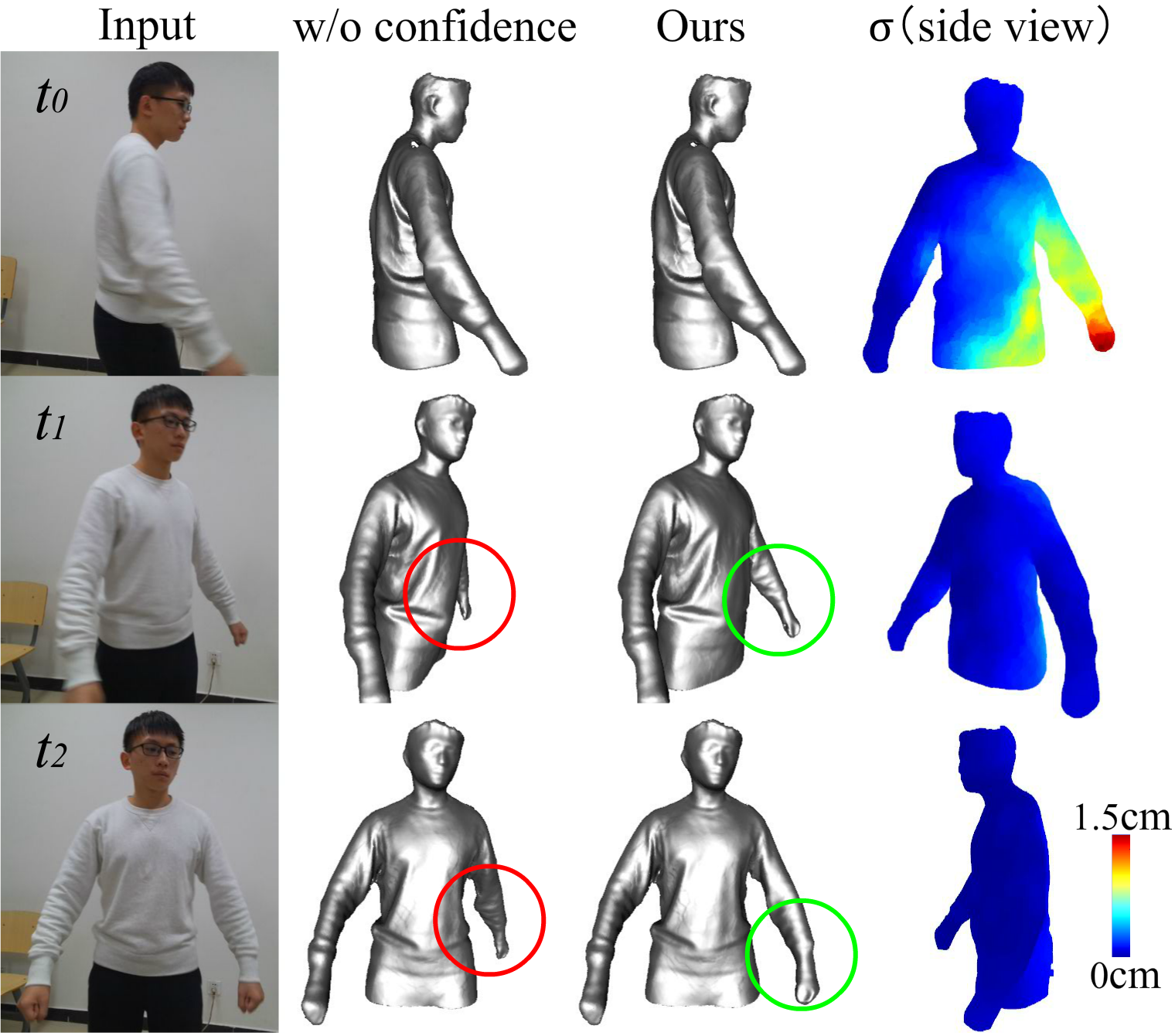}}
	\caption{Ablation study on confidence weights.}
	\label{fig:ablation_sigma}
\end{figure}

\begin{figure}[]
	\centerline{\includegraphics[width=0.90\linewidth]{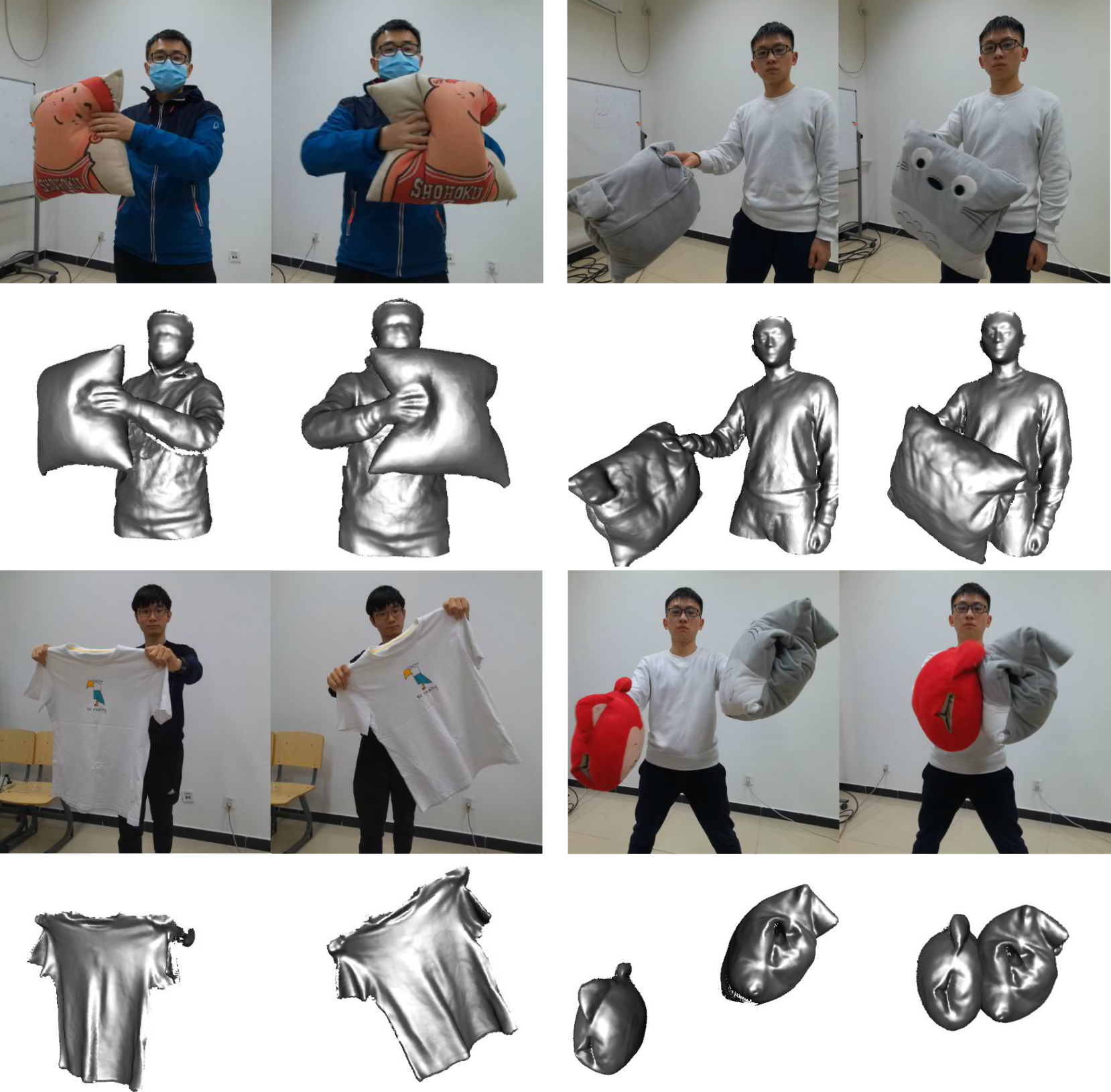}}
	\caption{More reconstruction results of our method.}
	\label{fig:more}
\end{figure}

\subsection{Experiments of Dynamic Reconstruction}
\label{sec:rec_eval}
We evaluate our method on the public non-rigid reconstruction benchmark of the DeepDeform~\cite{deepdeform} dataset. 
We first compare our method with the state-of-the-art of both online and offline solutions. 
Then we perform ablation studies to evaluate our key components.
Finally, we show more reconstruction results of our system.

\subsubsection{Comparisons}
The results are shown in Tab.~\ref{tab:deepdeform}. We can see that there is a big accuracy gap between the existing online~\cite{newcombe2015dynamicfusion, volumedeform} and offline~\cite{deepdeform, nnrt} methods, indicating it is difficult to improve the accuracy in the online scenario. 
With the help of the light-weight full motion estimation and confidence prediction, our method fills this gap and even outperforms the state-of-the-art offline methods~\cite{deepdeform, nnrt} with real-time performance.

As Guo et al.~\cite{guo2017real} did not evaluate errors on the DeepDeform~\cite{deepdeform} non-rigid reconstruction benchmark, we conduct a numerical comparison on a sequence from the DeepDeform dataset on our own. 
The geometry error of all the frames in the sequence is shown in Fig.~\ref{fig:geo_err}. 
Our approach keeps the geometry error low throughout the sequence, while Guo et al.~\cite{guo2017real} has difficulty tracking the fast motion in the second half of the sequence.

For qualitative evaluation, we show comparisons with online methods DynamicFusion~\cite{newcombe2015dynamicfusion} and Guo et al.~\cite{guo2017real} in Fig.~\ref{fig:cmp_online} and offline methods DeepDeform~\cite{deepdeform} and Bozic et al.~\cite{nnrt} in Fig.~\ref{fig:cmp_offline}. 
Our method handles faster motion much better than the online methods and aligns better with the input images than the offline methods. 

\subsubsection{Ablation Studies}
\label{sec:ablation}
We evaluate three key components of our technique regarding their contributions to the 3D dynamic reconstruction.

\noindent\textbf{Complete Node Motion.}
We remove the complete node motion constrain $E_\mathrm{motion}$ and show qualitative reconstruction results in Fig.~\ref{fig:ablation_seq}.
In each group of the results, the reconstructed models at $t_0$ are almost the same. 
After a few frames, the reconstruction results with the $E_\mathrm{motion}$ term are significantly better than the unused ones. 
Specifically, the body parts occluded at $t_1$ fail to be tracked by the method without $E_\mathrm{motion}$ term. 
Moreover, the geometry error curve shown in Fig.~\ref{fig:geo_err} indicates that without the $E_\mathrm{motion}$ term, the geometry error rises frequently, which is majorly caused by occlusions.
For more comparison, the results in Tab.~\ref{tab:deepdeform} show that removing $E_\mathrm{motion}$ term leads to higher reconstruction error.
These comparisons verify that the 3D full motion helps the reconstruction system track the occluded regions. 

\noindent\textbf{Temporal Information.}
We investigate the effectiveness of the temporal information used in the graph neural network, and show the comparisons in Fig.~\ref{fig:ablation_seq}. 
The body parts occluded at $t_1$ are not well tracked without the temporal information. 
When a moving arm is largely occluded, with only the motion of the visible upper arm, it is still hard to estimate the motion accurately. 
Involving the historical motion helps to improve the motion prediction.
Besides, results in Fig.~\ref{fig:geo_err} show that removing the temporal information also brings higher geometry error.

\noindent\textbf{Per Node Confidence.}
To show the importance of the per node confidence, we visualize the predicted standard deviation $\sigma$ on the reconstructed object's surface in the side view (from right to left) and the corresponding results in Fig.~\ref{fig:ablation_sigma}. 
At $t_0$, the left side of the body is occluded, and the predicted $\sigma$ of this region is significantly higher than that of the visible region, which indicates the corresponding motion is more uncertain.
As the left side of the body keeps occluded in a few frames, the inaccurate motion causes error to accumulate over time in the results without considering the confidence.
The method with per node confidence down-weights the untrustworthy motion and brings better reconstruction results.

\subsubsection{More Results}
We show more reconstruction results of our method in Fig.~\ref{fig:more}. Ours reconstruction system is able to handle various types of targets because the neural network is trained on a dataset that includes a rich set of non-rigid objects.

%% file: sec/5-limitations.tex
\section{Limitations}
Although our method improves the quality of motion tracking by introducing a graph-based full motion prediction network, there are still some failure cases. 
First, our method cannot handle topology change, which is an open problem for node-graph-based reconstruction systems. 
Incorrect connections in the deformation graph can lead to tracking failures. 
Possible solutions could be leveraging the advantage of tracking-free reconstruction methods like POSEFusion\cite{posefusion}, or using neural networks to predict the positions and connections of the graph nodes like Bozic et al.\cite{ndg}.
Besides, if a part of the object is occluded for a long time and has large non-rigid motion, it is hard for our method to re-track it when it becomes visible again.

%% file: sec/6-conclusion.tex
\section{Conclusion}
We propose OcclusionFusion, a single-view RGB-D based real-time dynamic 3D reconstruction method that outperforms the current online techniques by a large margin.
We use a complete 3D motion field to guide the object reconstruction and tracking, where the motion of the occluded regions is estimated online by a pre-trained light-weight graph neural network. 
The graph neural network combines the motion of visible regions and temporal information by involving an LSTM structure to accurately and efficiently predict the complete object motion.
Moreover, the graph neural network predicts confidence together with the motion by modeling the network output as a Gaussian distribution, which effectively enhances the robustness of the reconstruction system. 
Experimental results show that long and challenging sequences can be tracked well in real time using our technique with a single view input.

%% file: sec/7-appendix.tex
\begin{appendix}

\section{Network Details}
We first show the details in the network architecture, including the LSTM-based temporal motion encoding and the graph neural network.
Then, we show the details in the network training.
The neural network is implemented using the PyTorch\cite{torch} and PyTorch Geometric\cite{pyg} libraries.

\subsection{Architecture Details}
\noindent\textbf{LSTM-based Temporal Motion Encoding.}
The input and output motion of the LSTM module are represented using Gaussian distribution with diagonal covariance $\mathcal{N}(\boldsymbol{\mu}, \sigma^2\boldsymbol{I})$, where $\boldsymbol{\mu}\in \mathbb{R}^3, \sigma\in \mathbb{R}$.
The LSTM module is a standard two-layer LSTM with a hidden feature dimension of 32.
Then we use a fully connected layer to predict the $\boldsymbol{\mu}'$ and $\sigma'$ based on the output feature of the LSTM.

\begin{figure*}[]
	\centerline{\includegraphics[width=0.80\linewidth]{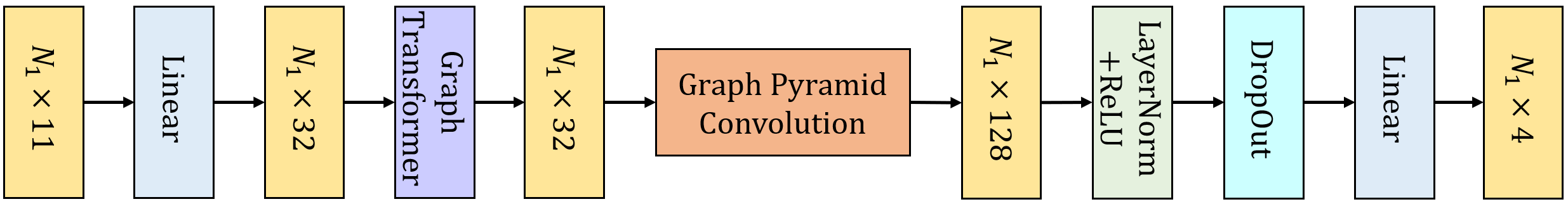}}
	\caption{Architecture of the graph neural network.}
	\label{fig:total}
\end{figure*}

\begin{figure*}[]
	\centerline{\includegraphics[width=1.0\linewidth]{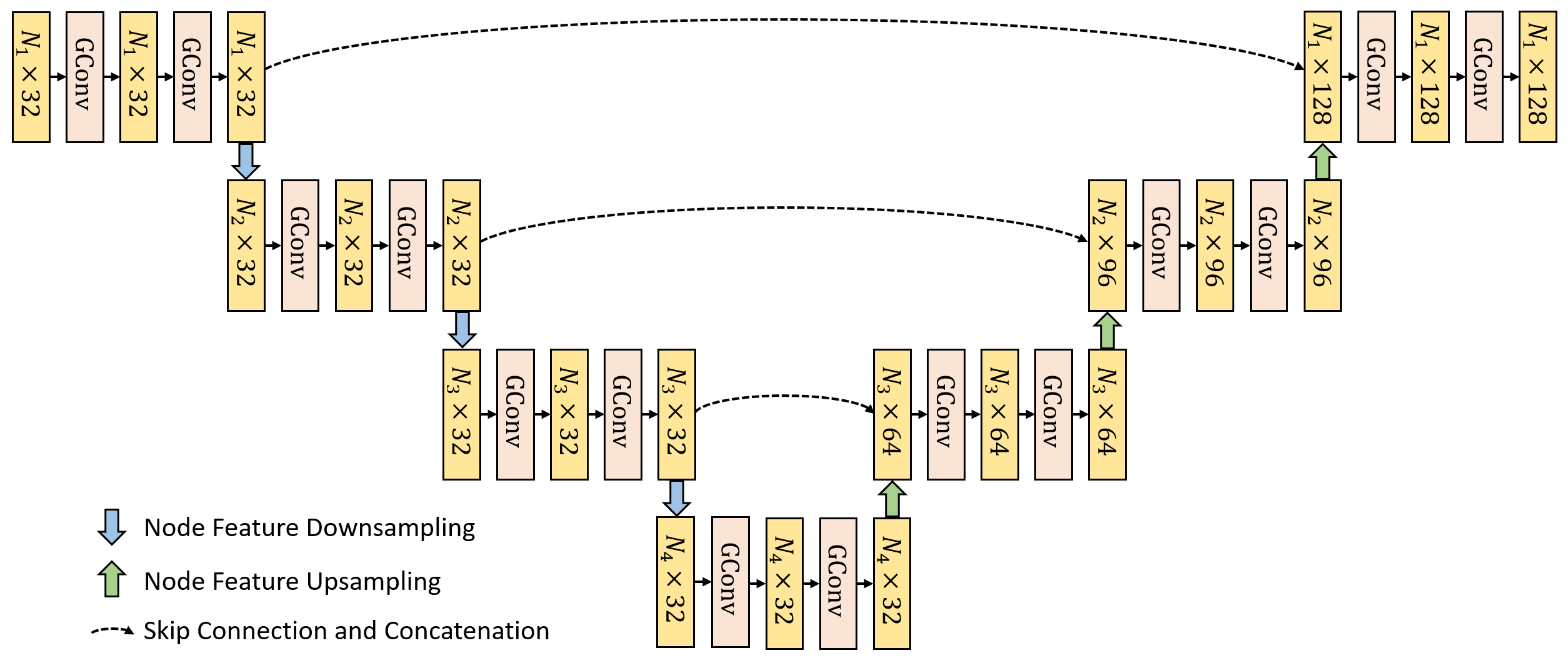}}
	\caption{Architecture of the Graph Pyramid Convolution  in Fig.~\ref{fig:total}.}
	\label{fig:pyd}
\end{figure*}

\begin{figure}[]
	\centerline{\includegraphics[width=0.25\linewidth]{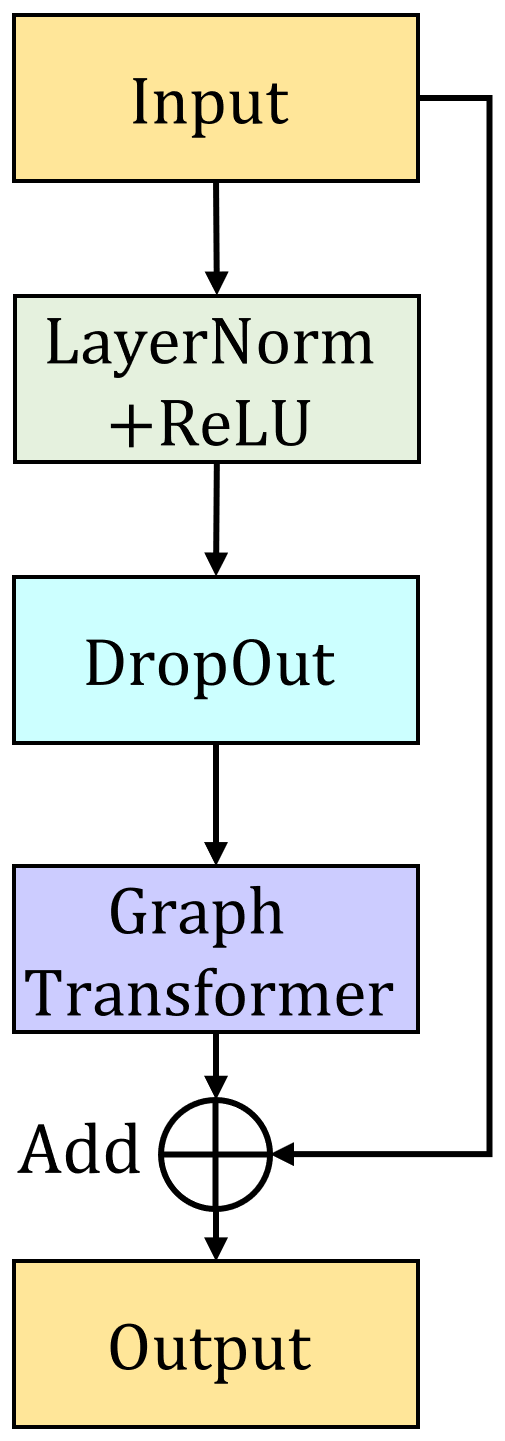}}
	\caption{Architecture of the GConv block.}
	\label{fig:block}
\end{figure}

\noindent\textbf{Graph Neural Network.}
We show the architecture of the graph neural network in Fig.~\ref{fig:total} and the architectures of the graph pyramid convolution and the GConv block in Fig.~\ref{fig:pyd} and Fig.~\ref{fig:block}.
The graph transformer is proposed by Shi et al.\cite{transformer_conv}. 
For all the dropout blocks, the drop probability is $0.1$.
$N_i$ denotes the number of nodes in the $i$th level of the graph pyramid. 
The dimensionality of the input node feature is $11$. 
It contains three dimensions of node position, three dimensions of node motion of the current frame, one dimension of the visibility, and four dimensions of the output $\boldsymbol{\mu}'$ and $\sigma'$ from the LSTM module. 
Then the output motion vectors ($\boldsymbol{\mu}$ and $\sigma$) of the graph neural network will be used as the input historical motion vectors for future frames.

\subsection{Training Details}
\noindent\textbf{Loss Function.}
The log-likelihood loss based on per node Gaussian distribution can be transformed as follow:
\begin{equation}
\begin{aligned}
    \mathcal{L} & = -\frac{1}{N}\sum_{i=1}^N \log \left(\mathcal{N}\left(\boldsymbol{y}_i \mid \boldsymbol{\mu}_{i}, \sigma_{i}^2 \boldsymbol{I}\right)\right) \\ & = \frac{1}{2N}\sum_{i=1}^N \left(\log{2\pi} + \log{\sigma_i} + \frac{\|\boldsymbol{y}_i - \boldsymbol{\mu}_i\|_2^2}{\sigma_i^2}\right)
    \\ & = C_1\sum_{i=1}^N \left(\log{\sigma_i} + \frac{\|\boldsymbol{y}_i - \boldsymbol{\mu}_i\|_2^2}{\sigma_i^2}\right) + C_2
    , 
\end{aligned}
\end{equation}
where $C_1, C_2$ are constants. Therefore the loss function can be simplified as:
\begin{equation}
    \mathcal{L} = \frac{1}{N} \sum_{i=1}^N \left(\log{\sigma_i} + \frac{\|\boldsymbol{y}_i - \boldsymbol{\mu}_i\|_2^2}{\sigma_i^2}\right).
\end{equation}
Thus, if the $\sigma_i$ is a fixed value, the loss is equivalent to the mean-squared error (MSE) loss.

\noindent\textbf{Training Procedure.}
The whole network is trained end-to-end. 
The LSTM module takes the historically predicted motion as input, while the predicted motion is not accurate enough at the beginning. 
So we use the ground truth historical motion with $\sigma=0$ to warm up the network training.
We first train the network using ground truth historical motion for 200 epochs. 
Then, we switch the input historical motion to the network's output and train for another 1200 epochs. 
Besides, the input historical motion vectors are treated as undifferentiable constants and detached from the computation graph.
Thus the gradients of the LSTM module do not flow back to the graph neural network.
We train the network using the Adam optimizer with a learning rate of 0.001 and a batch size of 64.
The whole training process takes about four days using an NVIDIA RTX GeForce 2080Ti GPU.

To quantitatively evaluate the generalization ability of our network, we train the model on the humanoids subset and test it on the animal subset for motion estimation in Sec.~\ref{sec:motion_eval}.
For other experiments, we use both subsets for training to achieve better reconstruction results. 

In addition, as the estimated 3D motion of the visible part in the real world is not as perfect as the synthetic dataset, we add random Gaussian noise to the visible motion during network training. 
The noise is represented as $\mathcal{N}(\boldsymbol{\mu}, \sigma^2\boldsymbol{I})$, where $\boldsymbol{\mu} = [0, 0, 0]$ and $\sigma$ is sampled from a uniform distribution $\mathcal{U}(-0.4, 0.4)$.
However, in the motion prediction evaluation in Sec.~\ref{sec:motion_eval} we do not add any noise.

\section{Details in Multi-scale Graph Pyramid Construction}
We construct a 4-level multi-scale graph for message passing among nodes. 
Considering that the node connectivity of the first level is computed based on the Euclidean distance, which may lead to misconnections between unrelated parts, we discard edges by temporal consistency. 
More precisely, if the distance between two nodes changes more than a threshold, the edge between them will be discarded.

The node features downsampling between adjacent levels is performed directly by copying the node features to the higher level, as the nodes in the $(l+1)$th level are a subset of the nodes in the $l$th level.
In feature upsampling, the node features of the $(l+1)$th level are assigned using the feature of the nearest neighbor in the $l$th level.

The intervals between nodes from the first to the fourth level are set to $\{ 4\mathrm{cm}, 8\mathrm{cm}, 16\mathrm{cm}, 32\mathrm{cm} \}$ and the neighbor amounts are $\{ 8, 6, 4, 3 \}$. Besides, the distance change threshold is set to $4\mathrm{cm}$.

\section{RGB-D Based Optical Flow Prediction}
We use the RAFT\cite{raft} network to estimate 2D optical flow. 
The original RAFT is trained with RGB images. 
We change the input dimensionality from 3 to 4 and retrain the network using RGB-D images as input. 
For the depth channel, we use the inverse depth (the reciprocal of depth) as input.

To speed up the 2D optical flow estimation in the 3D reconstruction system, we resize the input images from $640\times 480$ to $320\times 240$ to compute the optical flow and upsample the optical flow to the original resolution by bilinear interpolation.

\begin{figure}[]
	\centerline{\includegraphics[width=1.0\linewidth]{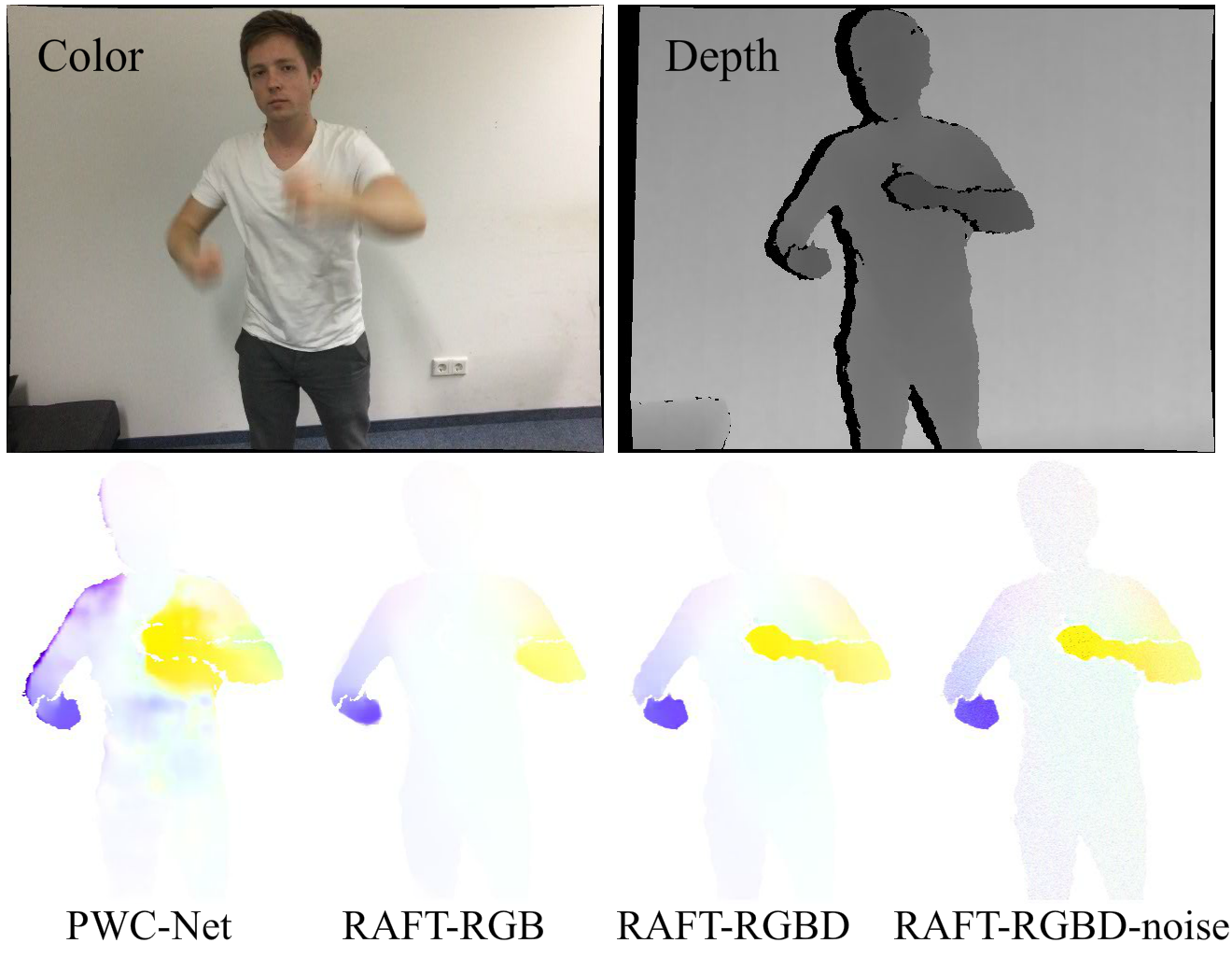}}
	\caption{Results of different optical flow methods when motion blur occurs.}
	\label{fig:flow}
\end{figure}

\begin{figure}[]
	\centerline{\includegraphics[width=0.9\linewidth]{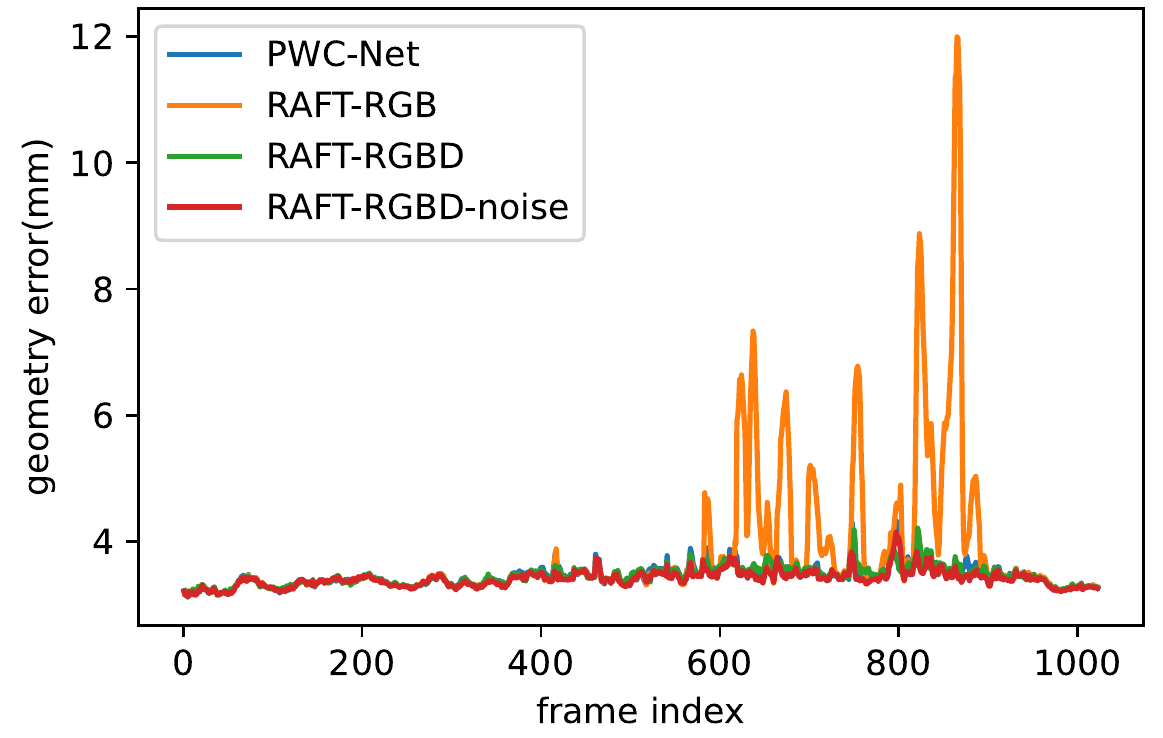}}
	\caption{Geometry errors on the top sequence of Fig.~\ref{fig:ablation_seq}. The average geometry errors over the whole sequence are $3.44\mathrm{mm}$, $3.79\mathrm{mm}$, $3.45\mathrm{mm}$, and $3.41\mathrm{mm}$ from PWC-Net to RAFT-RGBD-noise.}
	\label{fig:geo_err_flow}
\end{figure}

\begin{figure}[]
	\centerline{\includegraphics[width=1.0\linewidth]{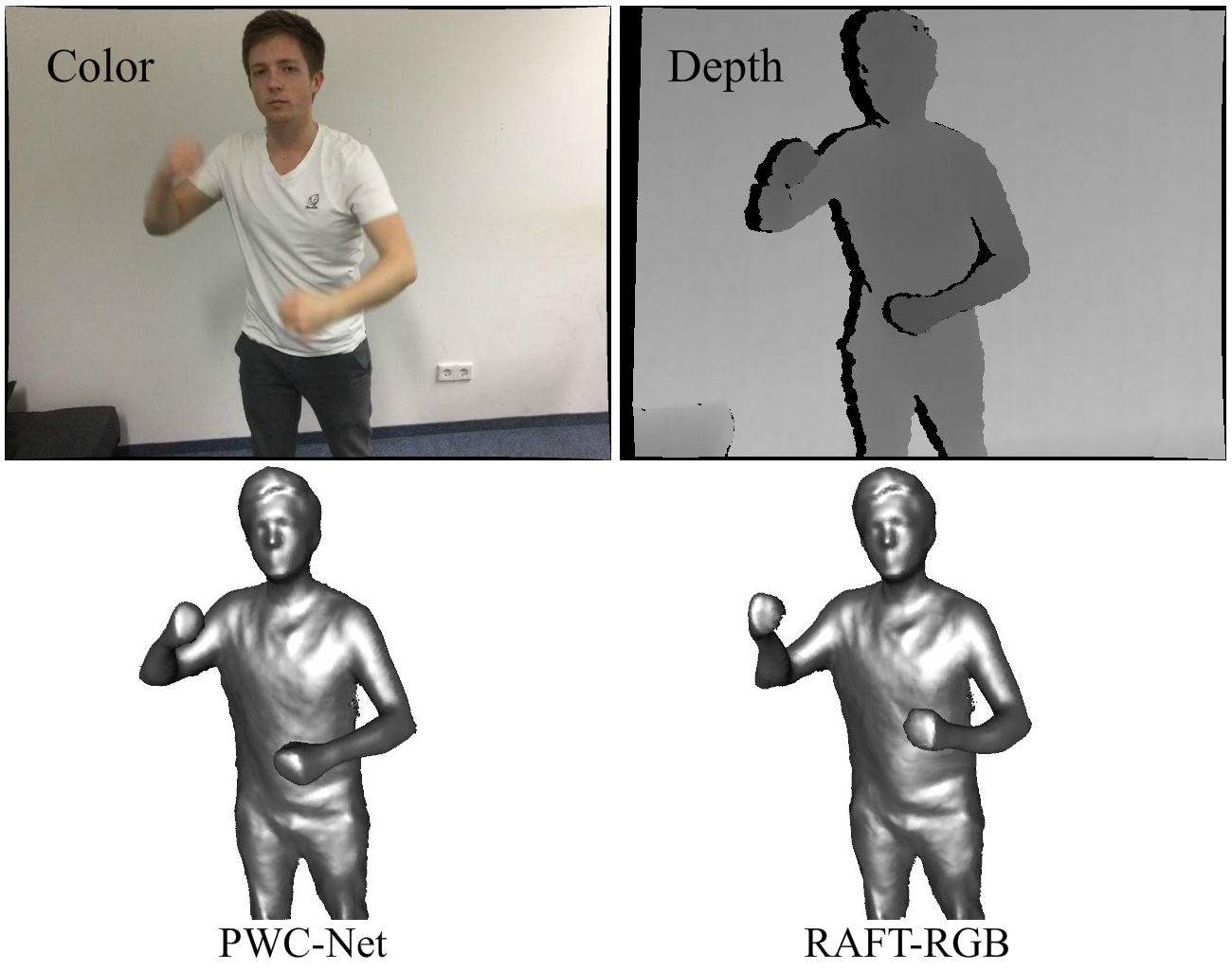}}
	\caption{Reconstruction results based on PWC-Net and RAFT-RGB.}
	\label{fig:flow_rec}
\end{figure}

\section{Choice of Optical Flow}
To test the robustness to flow estimation, we evaluate our system using different optical flow settings PWC-Net\cite{pwcnet}, RAFT-RGB and RAFT-RGBD.
Besides, we further add Gaussian noise of $\mathcal{N}(0, 4)$ pixels on $x$ and $y$ axes to our RAFT-RGBD optical flow.
We use the top sequence of Fig.~\ref{fig:ablation_seq} for evaluation.
The optical flow results at the 833rd frame of the sequence are shown in Fig.~\ref{fig:flow}. 
We can see severe motion blur occurs on the fast swinging arm in the color image, and significant errors appear in the optical flow of RGB-based methods (PWC-Net and RAFT-RGB).
However, since depth images do not suffer the blur artifacts much and provide geometric information, RAFT-RGBD generates reliable optical flow.
This indicates the benefit of involving depth in flow estimation.

For quantitative evaluation, we show the geometry errors of different optical flow methods over the whole sequence in Fig.~\ref{fig:geo_err_flow}.
We can see that the geometry errors of our reconstruction method based on PWC-Net, RAFT-RGBD, and RAFT-RGBD-noise are all low and close to each other, which indicates that our method is robust to noise in the optical flow.
Only the reconstruction result based on the RAFT-RGB optical flow provides a large geometry error, which is caused by the tracking failure of the fast swinging arm.
We show the reconstruction results using PWC-Net and RAFT-RGB at frame 837 (4 frames after the optical flow shown in Fig.~\ref{fig:flow}) in Fig.~\ref{fig:flow_rec}. 
The reconstruction result of PWC-Net is better than RAFT-RGB, although they both provide a larger optical flow error.
We believe the reason for this is that the optical flow errors of PWC-Net appear mainly on the torso, while those of RAFT-RGB appear on the anterior segment of the arm.
In addition, errors on the torso are more easily  corrected by the learned motion prior and the regularization term in the optimization, because there is more motion information around the torso region where the predicted optical flow is wrong.

\end{appendix}